\documentclass[letterpaper, 10 pt, journal, twoside]{IEEEtran}

\IEEEoverridecommandlockouts                              

\usepackage{graphicx}
\usepackage{amsmath}
\usepackage{hyperref}
\usepackage{comment}

\usepackage{color}

\renewcommand{\baselinestretch}{.97}

\begin{document}

\title{Re$^3$: \textit{Re}al-Time \textit{Re}current \textit{Re}gression Networks for \\ Visual Tracking of Generic Objects}

\author{Daniel Gordon$^{1}$ Ali Farhadi$^{1,2}$ and Dieter Fox$^{1}$%
\thanks{Manuscript received: September 5, 2017; Revised December 11, 2017; Accepted December 12, 2017.}
\thanks{This paper was recommended for publication by Editor Antonio Bicchi upon evaluation of the Associate Editor and Reviewers' comments.}%
\thanks{This work was funded in part by the National Science Foundation under contract number NSF-NRI-1525251, NSF-IIS-1338054, NSF-NRI-1637479, NSF-IIS-1652052, ONR N00014-13-1-0720, the Intel Science and Technology Center for Pervasive Computing (ISTC-PC), a Siemens grant, the Allen Distinguished Investigator Award, and the Allen Institute for Artificial Intelligence. We would also like to thank NVIDIA for generously providing a DGX used for this research via the UW NVIDIA AI Lab (NVAIL).}%
\thanks{\footnotesize $^{1}$Paul G. Allen School of Computer Science, University of Washington}
\thanks{\footnotesize $^{2}$Allen Institute for AI}
\thanks{{\tt\footnotesize danielgordon@cs.washington.edu}}
}

\markboth{IEEE Robotics and Automation Letters. Preprint Version. Accepted December, 2017}
{Gordon \MakeLowercase{\textit{et al.}}: Re3: Real-Time Recurrent Regression Networks}  

\maketitle

\begin{abstract}
   Robust object tracking requires knowledge and understanding of the object being tracked: its appearance, its motion, and how it changes over time. A tracker must be able to modify its underlying model and adapt to new observations. We present Re$^3$, a real-time deep object tracker capable of incorporating temporal information into its model. Rather than focusing on a limited set of objects or training a model at test-time to track a specific instance, we pretrain our generic tracker on a large variety of objects and efficiently update on the fly; Re$^3$ simultaneously tracks and updates the appearance model with a single forward pass. This lightweight model is capable of tracking objects at 150 FPS, while attaining competitive results on challenging benchmarks. We also show that our method handles temporary occlusion better than other comparable trackers using experiments that directly measure performance on sequences with occlusion.
   \end{abstract}



\section{Introduction}
  
    \IEEEPARstart{O}{bject} tracking plays an important role in many robotics applications. The main focus in the robotics community has been on developing trackers for known object types or specific object instances, such as boxes, hands, people, and cars, using RGB images or 2D/3D range data such as laser scans and depth images~\cite{arras2007using,premebida2007lidar,dart}.  This setting has the advantage that object-specific trackers can be designed or trained offline and that shape models of the objects are often available~~\cite{dart}.  However, in many scenarios it is not feasible to pre-specify what kind of objects needs to be tracked.  Examples include drone-based surveillance where a remote user specifies an object of interest by clicking on a single image frame~\cite{LanRSS17}, or learning from demonstration where a user picks up an unspecified object and the robot has to keep track of the object as a task is being demonstrated. In such settings, a robot must be able to quickly generate an internal model of the relevant object and continuously update this model to represent changes in the object's pose, shape, scale, and appearance, while being robust to appearance change due to external factors like occlusions and changes in lighting conditions. 
    
    Instead of assuming a known object model, we focus on the problem of \emph{generic} object tracking in RGB video data, which can be concisely phrased as: given a bounding box around an arbitrary object at time $t$, produce bounding boxes for the object in all future frames~\cite{vot2014}. In this paper, we only consider trackers which operate on streaming data; trackers cannot modify previous estimates given current or future observations. This requirement is necessary for many robotics settings, where a tracker is typically used in conjunction with another algorithm such as a reactive trajectory planner.
    
    Current generic 2D image tracking systems predominantly rely on learning a tracker online. A popular paradigm for tracking algorithms is tracking-by-detection: training an object-specific detector, and updating it with the object's new appearance at every frame. A disadvantage of this technique is that updating the tracker often takes a significant amount of time and computational resources. Conversely, object-specific trackers such as~\cite{premebida2007lidar} train detectors offline, but only function on these few object types.

\begin{figure}
    \centering
    \resizebox{\columnwidth}{!}{
    \includegraphics[]{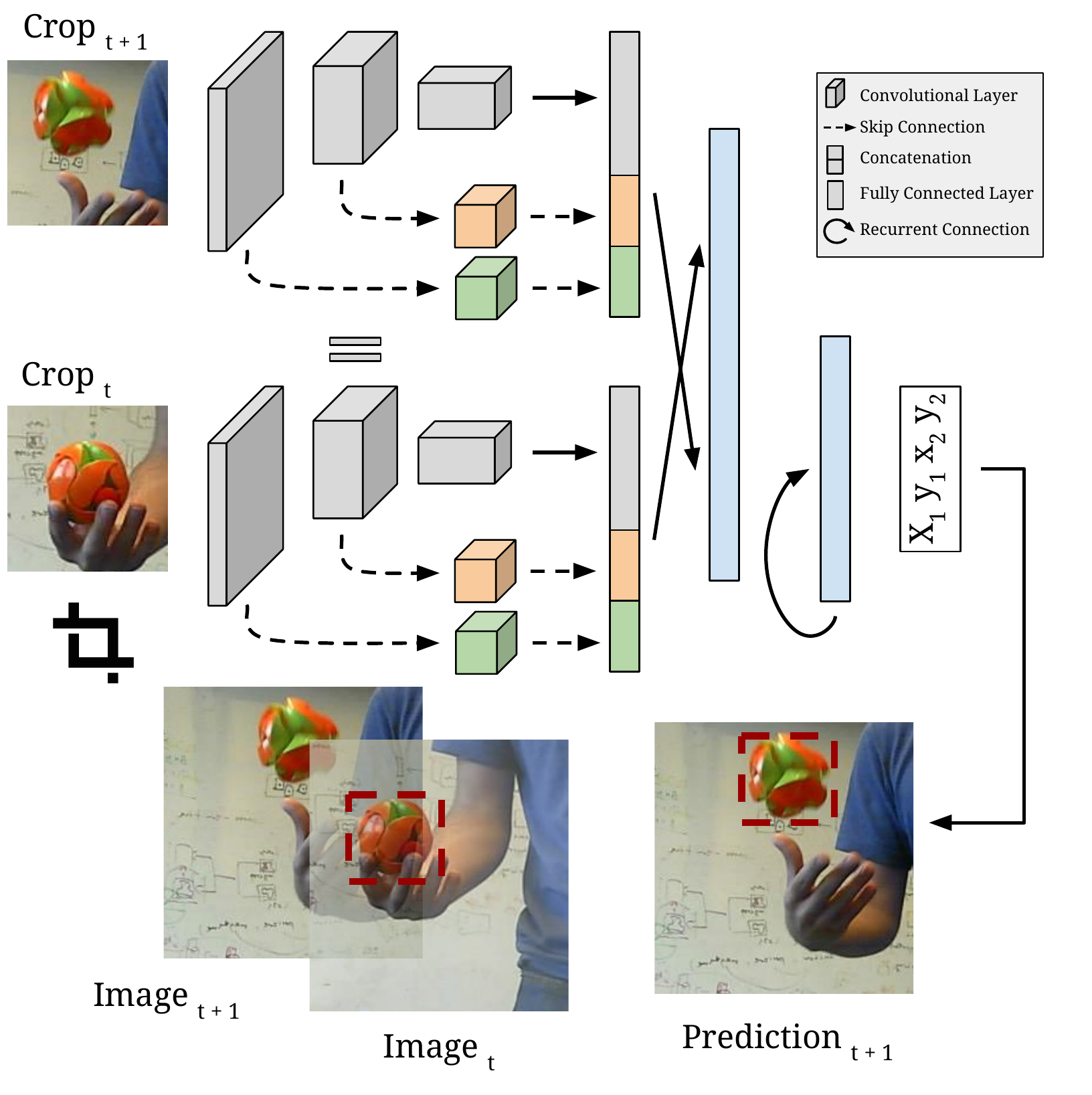}
    }\vspace*{-2ex}
    \caption{Network Structure: Image crop pairs are fed in at each timestep. Both crops are centered around the object's location in the previous frame, and padded to two times the width and height of the object. Before every pooling stage, we add a skip layer to preserve high-resolution spatial information. The weights from the two image streams are shared. The output from the convolutional layers feeds into a single fully connected layer and an LSTM. The network predicts the top left and bottom right corners of the new bounding box.}\vspace*{-4ex}
    \label{fig:network_layout}
\end{figure}

     We propose the \textit{Re}al-time, \textit{Re}current, \textit{Re}gression-based tracker, or Re$^3$: a fast yet accurate network for generic object tracking that addresses these disadvantages. Prior work has shown that given enough examples, a pretrained deep neural network can learn a robust tracker that functions on previously unseen objects~\cite{bertinetto2016fully,held}. However, instead of freezing the network as in~\cite{bertinetto2016fully,held} or adjusting the network parameters via online training~\cite{nam2016mdnet}, Re$^3$ learns to store and modify relevant object information in the recurrent parameters. By overcoming the need for any re-training, Re$^3$ efficiently tracks and updates itself simultaneously.
    
    By incorporating information from large collections of images and videos, our network learns to produce representations that capture the important features of the tracked object. The goal of this process is to teach the network how any given object is likely to change over time so these transformations can be embedded directly into the network. This shifts the computational burden offline, making Re$^3$ extremely fast and computationally cheap during inference, an important quality for algorithms operating on mobile robots with limited processing power. Because of our large variety of training data, we found our pretrained network can be directly applied to a variety of new environments such as drone videos, cellphone videos, and robot-mounted platforms, and due to the low computational cost, Re$^3$ could be run on embedded systems while remaining real-time. Our results show that recurrent networks are well suited for object tracking, as they can be fast, accurate, and robust to occlusions. Re$^3$ achieves competitive results on multiple tracking benchmarks, showing especially good performance during occlusions, all while running at 150 frames per second.

\section{Related Work}
    Object tracking has been studied in great depth by the robotics and computer vision community. In many cases, systems target objects with known 3D models or objects of a limited set of classes. DART~\cite{dart} requires a depth camera and a predefined articulated model, but produces fine-grained pixelwise labels. Ondruska et al.~\cite{ondruska2016deep} use planar laser scans and a recurrent network to track people under heavy occlusions. Their method succeeds because priors on likely human trajectories are quite strong. KITTI~\cite{kitti}, a popular vision and robotics benchmark suite, only tests performance on tracking cars and people. We focus on the harder problem of tracking arbitrary objects given only an initial bounding box. Generic object tracking represents a new challenge for convolutional neural networks. Most deep learning algorithms rely on having millions of examples to function, learning invariance to high-level concepts; object detection algorithms expect the network to learn what a person looks like, but not to differentiate between two people. Trackers, on the other hand, are often given only a single initial example and must specialize in order to track that specific target object. Because of the difficulty of adapting traditional deep methods to tracking, deep learning has only recently started to be used in tracking algorithms. In 2015, MDNet~\cite{nam2016mdnet}, a deep method, won the The Visual Object Tracking challenge (VOT)~\cite{vot2015} for the first time. The VOT reports~\cite{vot2014, vot2015, vot2016} present a succinct overview of many other generic object trackers. Those most related to ours can be categorized into three sub-groups: online-trained, offline-trained, and hybrid trackers.
    
    \noindent \textbf{Online-trained trackers:} The most prevalent type of trackers operate entirely online, continually learning features of the object of interest as new frames arrive. This includes keypoint-based and part-based trackers~\cite{edoardo2014matrioska}, correlation based methods~\cite{kcf}, and direct classification methods~\cite{hare2016struck}. These methods often rapidly train a classifier to differentiate between the object of interest, the background, and possible occluders. Discriminative Scale Space Tracker (DSST)~\cite{danelljan2017discriminative}, the winner of the VOT 2014 challenge~\cite{vot2014}, uses this approach. DSST learns discriminative correlation filters for different scale and translation amounts. Because online trackers must train on frames as they arrive, they tend to directly trade off speed with model complexity. 
    
    \noindent \textbf{Offline-trained trackers:} The success of deep learning is often attributed in part to its ability to utilize massive amounts of training data better than other machine learning methods. Offline trackers such as~\cite{bertinetto2016fully} and~\cite{held} employ this technique to great success. Because they are trained entirely offline, the networks are fast to evaluate at test time, allowing both methods to operate at faster than real-time speeds. However, this underscores a large problem with offline trackers: they do not adapt to what they are seeing. Instead of incorporating information from an entire track, they learn a similarity function between pairs of frames. Held et al.~\cite{held} use only a single frame history, meaning any amount of occlusion will confuse the tracker. Bertinetto et al.~\cite{bertinetto2016fully} rely solely on the initial frame for appearance information and try to detect the object in all subsequent frames, meaning large appearance changes, even if gradual, would be difficult to track. T-CNN~\cite{kang2017t} focuses on the detection and tracking problem in video by finding temporally coherent object detections, but they do not adapt the model using visual information from prior detections. Offline trackers' capabilities are fundamentally limited because they cannot adapt to new information.
    
    \noindent \textbf{Hybrid trackers:} Hybrid trackers attempt to solve the problems with online and offline trackers by taking the best from both. MDNet, the winner of the VOT 2015 challenge, trained an image classification network offline, and then learned a per-object classifier online~\cite{nam2016mdnet}. Similar approaches were taken by other top competitors~\cite{danelljan2015convolutional}. Still, the complexity of their online training techniques limited their methods to taking seconds to process each frame.
         
    Our approach is a hybrid tracker, but prioritizes offline learning and limits online adaptation to recurrent state updates. Although we make this trade-off, our method is substantially different from purely offline trackers because we use information from previous frames to make future predictions. This lets us model temporal dependencies between sequential images and reason about occlusions. Other recurrent trackers such as~\cite{kahou2017ratm} and~\cite{gan2015first} use attention-based recurrent neural networks. These techniques have only been shown to work on simple datasets such as tracking MNIST digits. To our knowledge, we are the first to demonstrate successful tracking in natural videos using recurrent neural networks.

\section{Method}

    Our tracking pipeline, depicted in Figure \ref{fig:network_layout}, consists of convolutional layers to embed the object appearance, recurrent layers to remember appearance and motion information, and a regression layer to output the location of the object. We train this network on a combination of real videos and synthetic data. At test time, unlike MDNet~\cite{nam2016mdnet}, we do not update the network itself; we instead let the recurrent parameters represent the tracker state which can be updated with a single forward pass. In this way, the tracker learns to use new observations to update the appearance and motion models, but no extra computational cost is spent on online training.

\subsection{Object Appearance Embedding}
    The task of generic object tracking in video sequences starts with an initial bounding box around an object, with the goal of keeping track of that object for the remainder of the video. For each frame of the video, the tracker must locate the object as well as update its internal state so it can continue tracking in future frames. A primary subtask in this framework is translating raw pixels into a higher-level feature vector representation. Many object trackers, like~\cite{edoardo2014matrioska} rely on extracting appearance information from the object pixels using hand-crafted features. We choose to learn the feature extraction directly by using a convolutional pipeline that can be trained fully end-to-end on a large amount of data.
    
    \noindent \textbf{Network Inputs:} Similar to~\cite{held}, at each frame, we feed the network a pair of crops from the image sequence. The first crop is centered at the object's location in the previous image, whereas the second crop is in the \textit{same} location, but in the \textit{current} image. The crops are each padded to be twice the size of the object's bounding box to provide the network with context. This padding offers a reasonable trade-off between speed, resolution, and search region size. If the bounding box at frame $j$ had centers $(X^j_c, Y^j_c)$ and width and height $W^j, H^j$, both crops would be centered at $(X^j_c, Y^j_c)$ with width and height $2W^j$ and  $2H^j$. By feeding a pair of crops, the network can directly compare differences in the two frames and learn how motion affects the image pixels. Though this method does not guarantee the object to be in the crop, if our first crop was in the correct location, the object would have to move more than 1.5 times its width and height in a single frame to be fully out of the crop, which is quite unlikely. The crops are warped to be $227 \times 227$ pixels before being input into the network. We experimentally determined that preserving the aspect ratio of the source images hurts performance because it forces the network to directly regress the aspect ratio rather than regress changes to the ratio. The pair of image features are concatenated at the end of the convolutional pipeline (late fusion) rather than at the beginning to allow the network to fully separate out the differences between the two images.
        
    \noindent \textbf{Skip Connections:} The hierarchical structure of convolutional networks extracts different levels of information from different layers~\cite{zeiler2014visualizing}; the lowest layers of image classification networks output features like edge maps, whereas the deeper layers capture high-level concepts such as animal noses, eyes, and ears~\cite{zeiler2014visualizing}. Rather than only using the outputs from the last layer of the network, we represent the object's appearance using low, mid, and high level features. We use skip connections when spatial resolution decreases to give the network a richer appearance model. In this way, the network can differentiate a person (high level concept) wearing a red (low level concept) shirt from a person wearing a blue shirt.
    
    The skip connections are each fed through their own $1 \times 1 \times C$ convolutional layers where $C$ is chosen to be less than the number of input channels. This reduces the dimensionality of the layers with higher spatial resolutions to keep computational cost low. As the spatial resolution is halved, $C$ is doubled. All skip connection outputs and the final output are concatenated together and fed through a final fully-connected layer to further reduce the dimensionality of the embedding space that feeds into the recurrent pipeline.

\subsection{Recurrent Specifications}
    Recurrent networks tend to be more difficult to train than typical feed-forward networks, often taking more iterations to converge and requiring more hyperparameter selection. We present a method of training a recurrent tracking network which translates the image embedding into an output bounding box while simultaneously updating the internal appearance and motion model. We also describe techniques that lead to faster convergence and better-performing networks.

    \noindent \textbf{Recurrent Structure:} Using the prior work of Greff et al.~\cite{greff2016lstm}, we opt for a two-layer, factored LSTM (the visual features are fed to both layers) with peephole connections.  We find that this outperforms a single layer LSTM even given a deeper convolutional network and larger embedding space. The two layer LSTM is likely able to capture more complex object transformations and remember longer term relationships than the single layer LSTM. The exact formulation is shown in Equations 1-6 where $t$ represents the frame index, $x^t$ is the current input vector, $y^{t-1}$ is the previous output (or recurrent) vector, $\mathbf{W}$, $\mathbf{R}$, and $\mathbf{P}$ are weight matrices for the input, recurrent, and peephole connections respectively, $b$ is the bias vector, $h$ is the hyperbolic tangent function, $\sigma$ is the sigmoid function, and $\odot$ is point-wise multiplication. A forward pass produces both an output vector $y^t$, which is used to regress the current coordinates, and the cell state $c_t$, which holds important memory information. Both $y_t$ and $c_t$ are fed into the following forward pass, allowing for information to propagate forward in time.

    \small
    \begin{align*}
        z^t = &~ h(\mathbf{W}_zx^t + \mathbf{R}_zy^{t-1} + b_z) &\text{LSTM input (1)} \\
        i^t = &~\sigma(\mathbf{W}_ix^t + \mathbf{R}_iy^{t-1} + \mathbf{P}_ic^{t-1} + b_i) &\text{input gate (2)} \\
        f^t = &~ \sigma(\mathbf{W}_fx^t + \mathbf{R}_fy^{t-1} + \mathbf{P}_fc^{t-1} + b_f) &\text{forget gate (3)} \\
        c^t = &~ i^t \odot z^t + f^t \odot c^{t-1} &\text{cell state (4)} \\
        o^t = &~ \sigma(\mathbf{W}_ox^t + \mathbf{R}_oy^{t-1} + \mathbf{P}_oc^t + b_o) &\text{output gate (5)} \\
        y^t = &~ o^t \odot h(c^t) &\text{LSTM output (6)}
    \end{align*}
    \normalsize
    
\begin{figure}
    \centering
    \vspace*{2ex}
    \resizebox{\columnwidth}{!}{
    \includegraphics[]{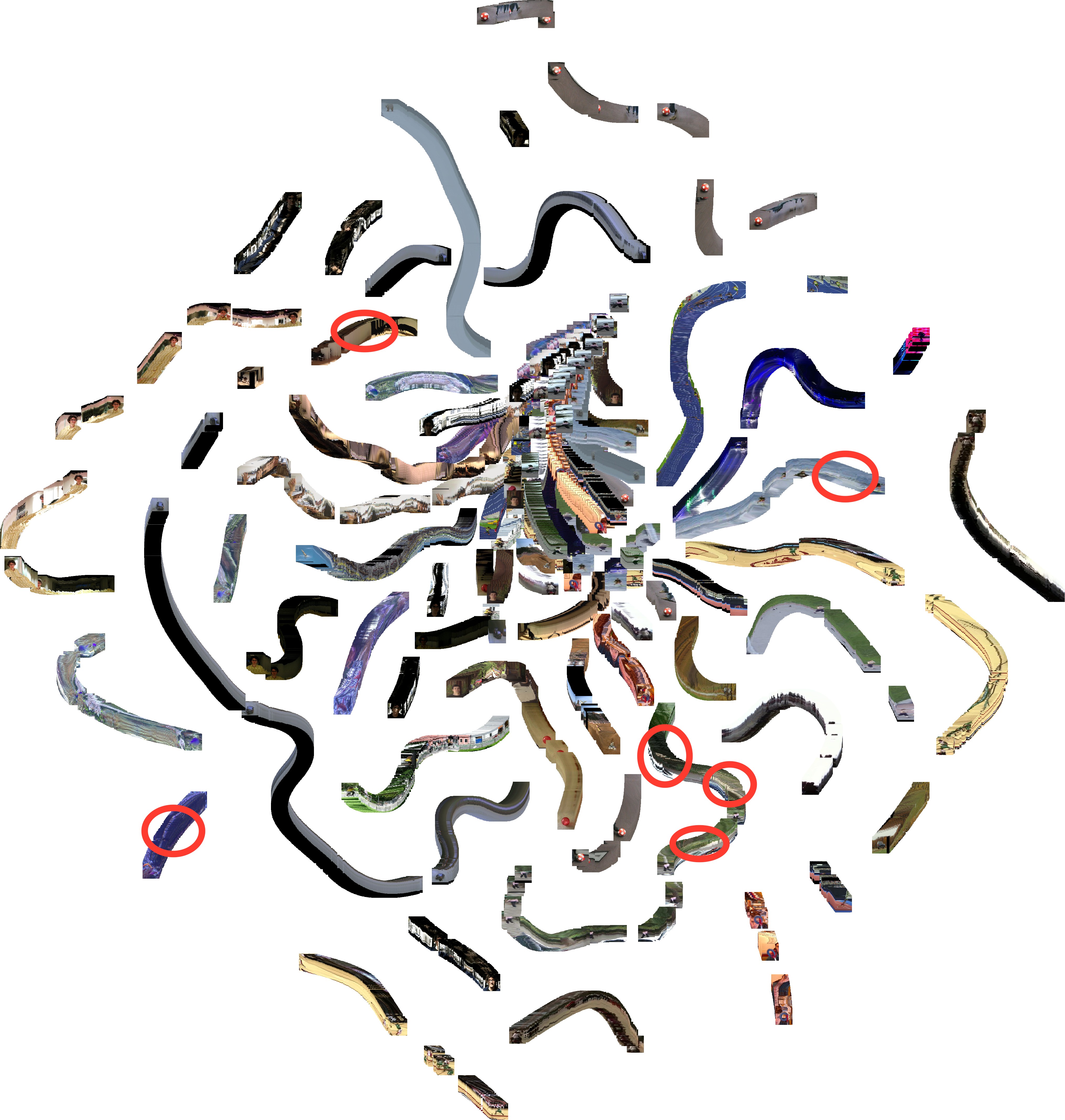}
    }\vspace*{-1ex}
    \caption{t-SNE embedding of LSTM states from VOT 2014~\cite{vot2014} data. The cell and output states are concatenated together to form the feature vector. Rather than forming clusters as is typical with t-SNE embeddings, the states form paths indicating that as the images change during a video, the LSTM states change in a similar fashion. Circled portions of the embedding indicate occlusion.}\vspace*{-5ex}
    \label{fig:tsne}        
\end{figure}

    The output and cell state vectors update as the object appearance changes. Figure \ref{fig:tsne} shows a t-SNE~\cite{tsne} plot of the LSTM states our tracker produces for each frame from the the VOT 2014~\cite{vot2014} videos. Because the LSTM states are initialized to 0 at the start of each video, the embeddings of the first few frames from each track are clustered together. As each video progresses, the LSTM state is transformed, resulting in many long, thin paths that follow the ordering of the frames in the original video. Certain points in the sequences with significant occlusion are circled, demonstrating that are embedding does not change drastically during occlusions even though the image pixels look quite different. The gaps in the sequences are mostly due to fast movement which tend to cause a rapid appearance change in the second crop.

    \noindent \textbf{Network Outputs:} The second LSTM's outputs are fed into a fully-connected layer with four output values representing the top left and bottom right corners of the object box in the crop coordinate frame, as is done in~\cite{held}. By regressing these coordinates, we can directly handle size and aspect ratio changes. Similar to~\cite{held}, we use an L1 loss on the outputs to encourage exact matches the ground truth and limit potential drift.
    
    \noindent \textbf{Unrolling during training:} Recurrent networks generally take many more iterations to converge than comparable feed-forward networks. This is likely because the inputs are all fed in sequentially, and then one or many outputs and losses are produced. This means the loss must propagate through many noisy intermediate states, causing the gradients to fluctuate and often not be useful for convergence. However, for tracking, each input is directly paired with an immediate output. Thus, we can use a training curriculum that begins with few unrolls, and slowly increases the time horizon that the network sees to teach it longer-term relationships. Without the shorter unroll step, the network may take exponentially longer to train, or may simply never converge. Specifically, we initially train the network with only two unrolls and a mini-batch size of 64. After the loss plateaus, we double the number of unrolls and halve the mini-batch size until a maximum unroll of 32 timesteps and a mini-batch size of 4. Using this curriculum, we do not find it necessary to clip gradients. 
    
    \noindent \textbf{Learning to Fix Mistakes:} Recurrent networks are often trained by feeding ground truth outputs into the future timesteps rather than the network's own predictions~\cite{fragkiadaki2015recurrent, vinyals2015show}. However, if we always provide the network with ground-truth crops, at test time it quickly accumulates more drift than it has ever encountered, and loses track of the object. To counteract this, we employ a regime that initially relies on ground-truth crops, but over time the network uses its own predictions to generate the next crops. We initially only use the ground truth crops, and as we double the number of unrolls, we increase the probability of using predicted crops to first $0.25$, then subsequently $0.5$ and $0.75$. This method is similar to the one proposed in~\cite{bengio2015scheduled}, however we make our random decision over the whole sequence rather than at every step independently.
    
\begin{figure*}
    \centering
    \resizebox{.9\textwidth}{!}{
    \includegraphics{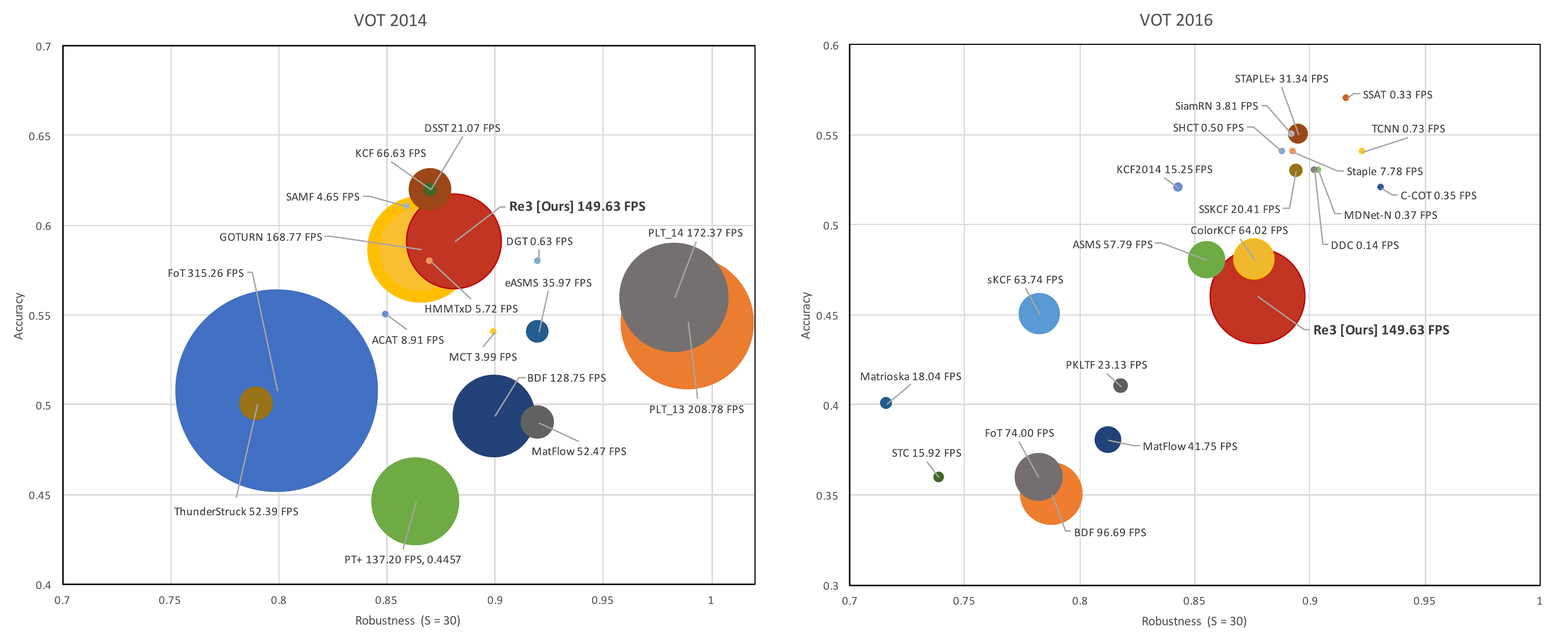}
    }\vspace*{-2ex}
    \caption{We compare Re$^3$ to other trackers on the VOT 2014~\cite{vot2014} and VOT 2016~\cite{vot2016} test suites. The size of the point indicates the speed of the tracker. Those below 3 FPS are enlarged to be visible. Speeds are taken directly from the VOT 2014~\cite{vot2014} and VOT 2016~\cite{vot2016} result reports. The VOT authors have stated that speed differences between years can be due to different code, different machines, and other confounding factors. For detailed analysis of other trackers' performance, please view~\cite{vot2014, vot2016}.}\vspace*{-4ex}
    \label{fig:speed_plot}
\end{figure*}

\subsection{Training Procedure}
    We use a combination of real and synthetic data to train our deep network. This results in our tracker being able to work on a large variety of object types, allowing us to successfully track across multiple datasets.
    
    \noindent \textbf{Training from Video Sequences:} We train Re$^3$ on two large object tracking datasets: the training set from the ILSVRC 2016 Object Detection from Video dataset (Imagenet Video)~\cite{imagenetVideo} and the Amsterdam Library of Ordinary Videos 300++ (ALOV)~\cite{alov}. In its training set alone, Imagenet Video provides 3862 training videos with 1,122,397 images, 1,731,913 object bounding boxes, and 7911 unique object tracks. This is by far the largest object tracking dataset we are aware of, however it only contains videos for 30 object categories. ALOV consists of 314 videos. We do not use the 7 videos that also occur in VOT 2014~\cite{vot2014} in order to avoid training on the test set. The remaining dataset comprises 307 videos and 148,319 images, each with a single object.
    
    \noindent \textbf{Training from Synthetic Sequences:} Recently, many deep methods have supplemented their training sets with simulated or synthetic data~\cite{held, richter2016playing}. Due to the large variety of objects labeled in `object detection in image' datasets, we construct synthetic videos from still images to show the network new types of objects. We use images from the Imagenet Object Detection dataset to fill this role~\cite{imagenetVideo}. We discard objects that are less than $0.1^2$ of the total image area due to lack of detail, resulting in 478,807 object patches.
    
    To generate simulated data, we randomly sample over all images for an object to track. We use random patches from the same image as occluder patches. The full image serves as the background for the scene. The object, background, and occluders are taken from the same image in order to keep our simulated images close to the real image manifold. We then simulate tracks for the object and occluders, at each timestep modifying an initial speed, direction, and aspect ratio with Gaussian noise. This data adds diversity to the types of objects that the network sees, as categories like ``person,'' which are common in many tracking datasets, are absent in Imagenet Video~\cite{imagenetVideo}.

    \noindent \textbf{Tracking at test time:} To generate test-time predictions, we feed crops to the network from each sequential frame. After every 32 iterations, we reset the LSTM state. This is necessary because we train on sequences with a maximum length of 32 frames, and without this reset, the LSTM parameters tend to diverge from values the network has seen before. Rather than resetting the LSTM state to all zeros, we use the output from the first forward pass. This maintains an encoding of the tracked object, while allowing us to test on sequences much longer than the number of training unrolls. We also notice that the reset helps the model recover from drifts by using a well-centered crop embedding.

    \noindent \textbf{Implementation Details:} We use Tensorflow~\cite{tensorflow} to train and test our networks \footnote{The Tensorflow code as well as pretrained network weights are available at \url{https://gitlab.cs.washington.edu/xkcd/re3-tensorflow}.}. Unless otherwise noted, we use the CaffeNet convolutional pipeline initialized with the CaffeNet pretrained weights for our convolutional layers. The skip connections occur after ``norm1,'' ``norm2,'' and ``conv5,'' with 16, 32, and 64 channels respectively. Each skip layer has a PReLU nonlinearity~\cite{prelu}. The embedding fully-connected layer has 2048 units, and the LSTM layers have 1024 units each. We initialize all new layers with the MSRA initialization method~\cite{prelu}. We use the the ADAM gradient optimizer~\cite{kingma2014adam} with the default momentum and weight decay and an initial learning rate of $10^{-5}$, which we decrease to $10^{-6}$ after 10,000 iterations and continue for approximately 200,000 iterations which takes roughly one week. All layers, including the pretrained ones, are updated with this learning rate. During training, we randomly mirror entire tracks with probability $0.5$. All tests were carried out using an Intel Xeon CPU E5-2696 v4 @ 2.20GHz and an Nvidia Titan X (Pascal). For timing purposes, we ignore disk read speeds as they are independent of the tracking algorithm used.
\section{Experiments}
    We compare Re$^3$ to other tracking methods on several popular tracking datasets in terms of both overall performance and robustness to occlusion. On all datasets, we are among the fastest, most accurate, and most robust trackers. We initially demonstrate our effectiveness by testing on a standard tracking benchmark, the Visual Object Tracking 2014 and 2016 (VOT 2014 and VOT 2016) challenges~\cite{vot2014, vot2016}, where we outperform other real-time trackers and are competitive with other deep methods. Next, we show results on the ILSVRC 2016 Object Detection from Video challenge (Imagenet Video)~\cite{imagenetVideo} comparing with other real-time trackers. We then examine our performance during occlusion with specific experiments on occluded data. Additionally, we perform an ablation study to understand the contributions of each part to the overall success of the method. Finally, we examine qualitative results on novel video domains such as drone footage and cellphone video.

\subsection{VOT 2014 and 2016}
    The VOT 2014 and 2016 object tracking test suite~\cite{vot2014, vot2016} consists of 25 and 60 videos respectively made with the explicit purpose of testing trackers. Many of the videos contain difficulties such as large appearance change, heavy occlusions, and camera motion. Trackers are compared in terms of accuracy (how well the predicted box matches with the ground truth) and robustness (how infrequently a tracker fails and is reset). More details about these criteria can be found in~\cite{vot2014}. Figure \ref{fig:speed_plot} compares Re$^3$ with other trackers submitted to the VOT 2014 and 2016 challenges~\cite{vot2014, vot2016} as well as with Held et al~\cite{held}. We show the 10 fastest trackers as well as the 10 most accurate trackers from each year.

    Figure \ref{fig:speed_plot} Left shows our full model trained using all of the available training data. We are among the most accurate methods overall, and among the most robust of the real-time methods, likely due to the LSTM's ability to directly model temporal changes, allowing the network to adapt without much computational overhead.
    
    Figure \ref{fig:speed_plot} Right compares our results against more modern trackers on the more difficult VOT 2016 test set~\cite{vot2016}. For training this model, we omit the ALOV data entirely since there is a large overlap between the two video sets. We later explore the detrimental effect this has on our network's performance in the ablation analysis (model H in Table \ref{table:ablation}). Re$^3$ is 450x faster than the best methods~\cite{ccot, vot2016}, while scoring only 20\% and 5\% lower in terms of relative accuracy and robustness. On both datasets, Re$^3$ offers an attractive trade-off of speed, accuracy, and robustness, especially in time-critical or computationally limited scenarios.

\subsection{Imagenet Video}
\begin{figure}
    \centering
    \vspace*{2ex}
    \resizebox{\columnwidth}{!}{
    \includegraphics[]{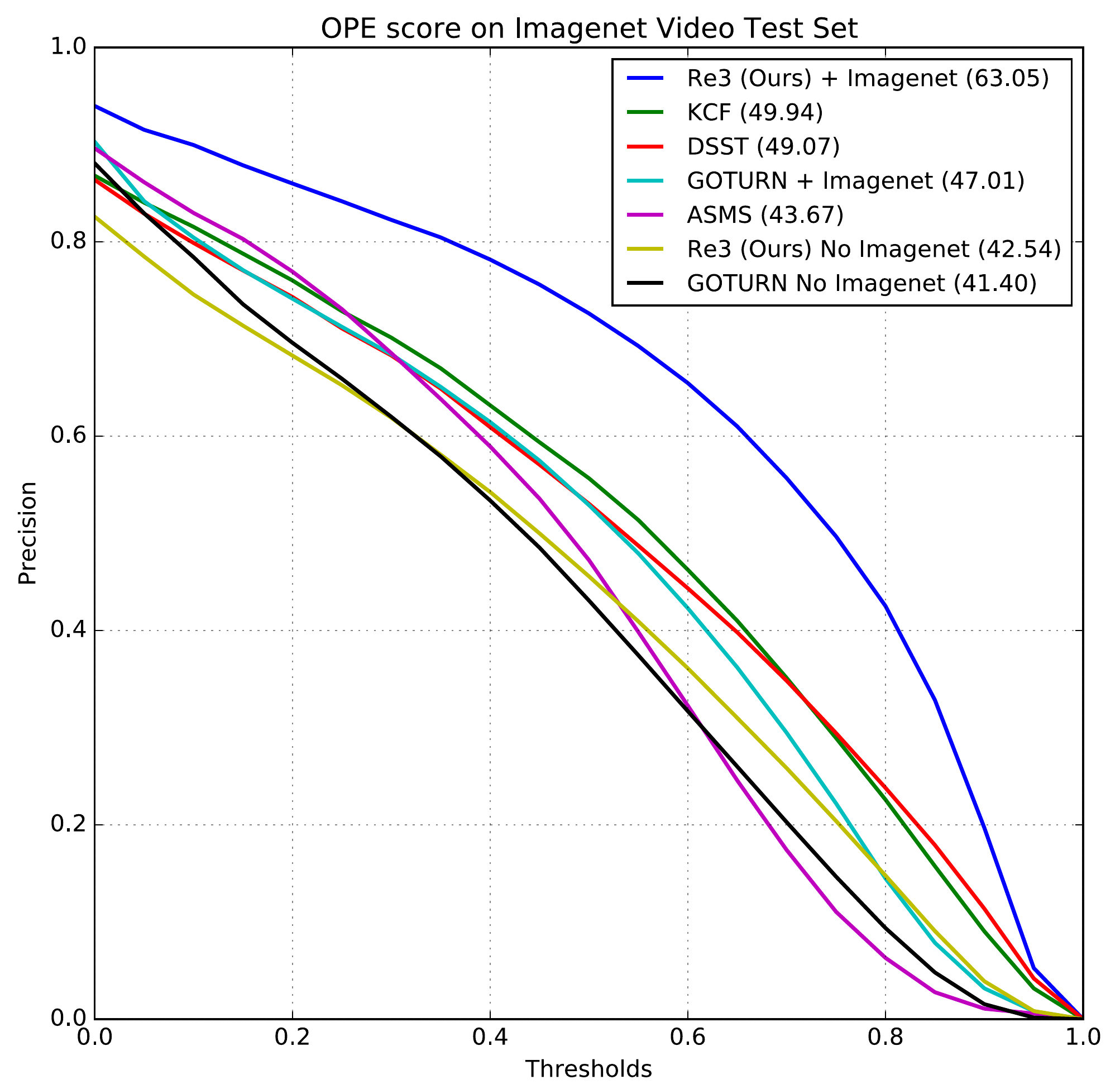}
    }\vspace*{-2ex}
    \caption{Various real-time trackers evaluated on the Imagenet Video test set~\cite{imagenetVideo}. Area under the curve (AUC) is shown for each method. We compare results with~\cite{kcf, dsst, held, asms} with code provided by~\cite{kcfCode, dsstCode, held, asmsCode} respectively.}\vspace*{-5ex}
    \label{fig:imagenet_plot}
\end{figure}

    The Imagenet Video validation set consists of 1309 individual tracks and 273,505 images~\cite{imagenetVideo}. It is the largest dataset we test on, and it offers significant insights into the success cases and failure cases of our tracker. We use Imagenet Video to evaluate our performance against other open-source real-time trackers. Each tracker is initialized with the first frame of each test sequence and is not reset upon losing track. Each individual bounding box is evaluated against the ground truth for that track at various IOU thresholds. Figure \ref{fig:imagenet_plot} shows our method outperforming other real-time trackers over all thresholds by a wide margin, though only our method and GOTURN~\cite{held} + Imagenet were trained with the Imagenet Video training set. We also train a version of our network without using the Imagenet Video training data, only using a combination of ALOV~\cite{alov} and simulated data. This performs significantly worse, most likely because LSTMs tend to take more data to train than comparable feed forward methods and this omits 90\% of our real training data. With sufficient training data, our method outperforms other methods trained on the same data.
    
\subsection{Online Object Tracking benchmark}
\begin{figure}
\resizebox{\columnwidth}{!}{
    \centering
    \includegraphics{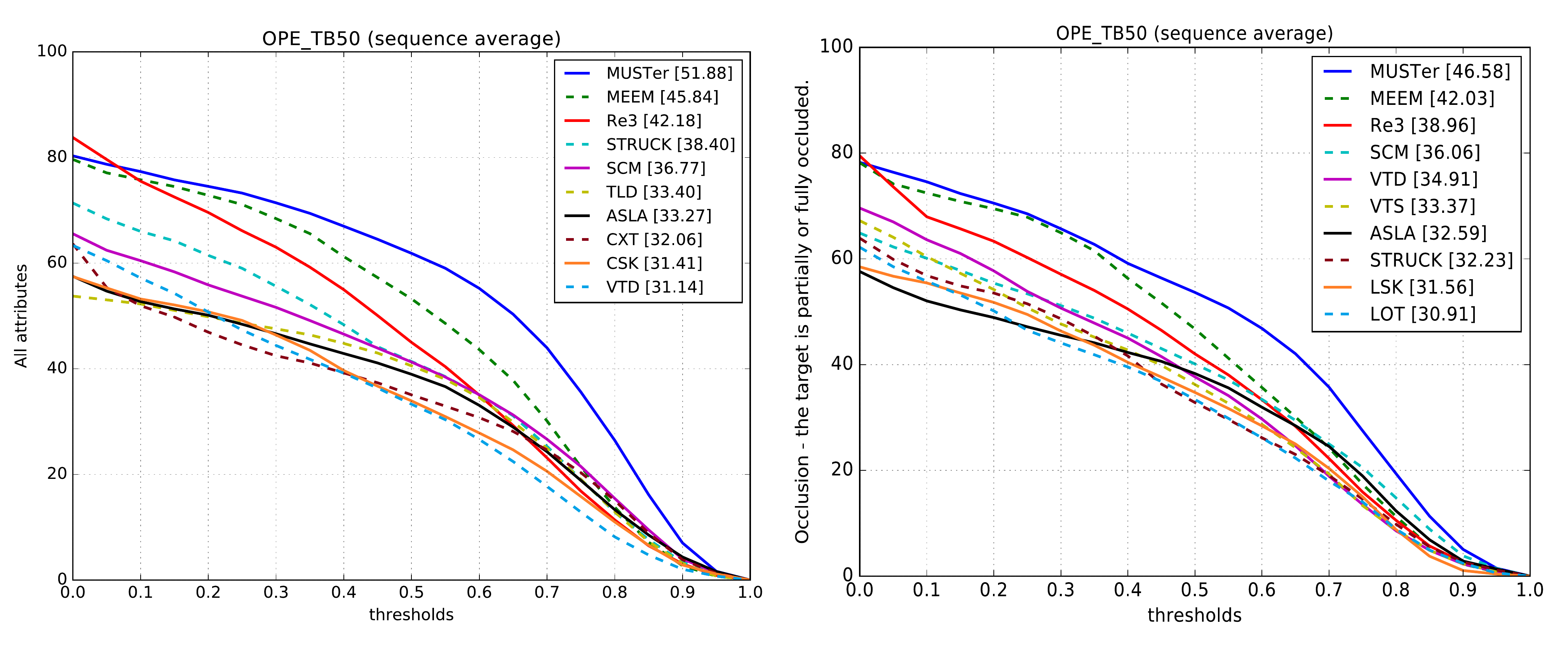}
    }
    \caption{Evaluation on the OTB benchmark~\cite{otb}. We examine performance both overall (left) and during occluded frames (right) using the One Pass Evaluation (OPE) criterion explained in~\cite{otb}. The legend shows area under the curve (AUC) for each method. Relative to other trackers, we suffer a smaller loss in accuracy due to occlusion. For detailed analysis of other trackers' performance, please view~\cite{otb}.}\vspace*{-4ex}
    \label{fig:ope_tb50}
\end{figure}

    The Online Object Tracking benchmark (OTB)~\cite{otb} is a widely used benchmark in tracking literature consisting of 50 challenging tracking videos of various objects. The One Pass Evaluation (OPE) criteria on OTB is equivalent to the evaluation we perform on Imagenet Video. In Figure \ref{fig:ope_tb50}, we show results competitive with the provided baselines from the OTB website~\cite{otb}, even though we again omit the ALOV training data.

\begin{figure}
    \centering
    \vspace*{2ex}
    \resizebox{\columnwidth}{!}{
    \includegraphics[]{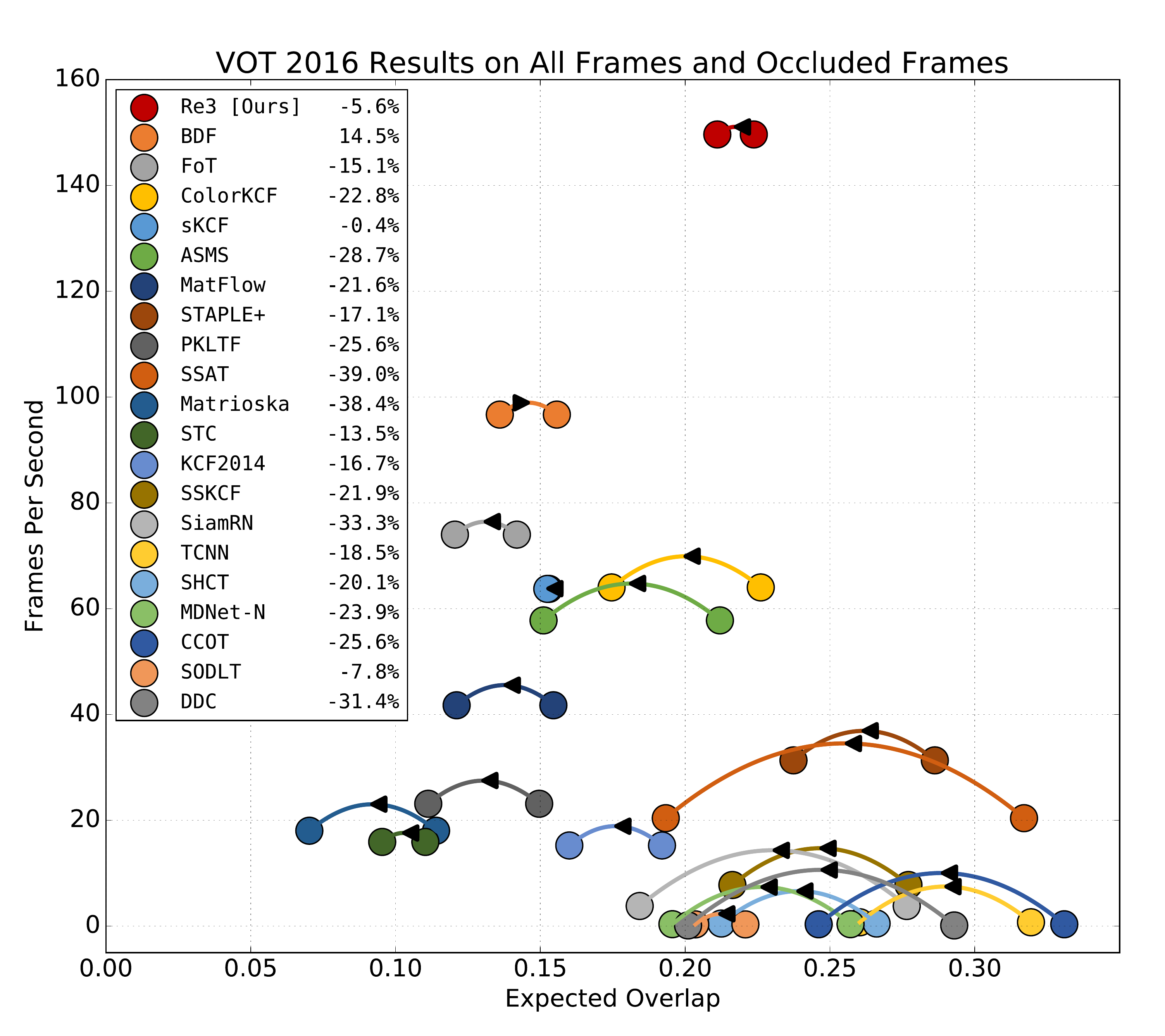}
    }\vspace*{-2ex}
    \caption{Expected overlap of various trackers compared between all frames and occluded frames. The arrow indicates that BDF improves during occlusion whereas all other methods degrade. For further analysis of compared trackers, please view~\cite{vot2016}.}\vspace*{-2ex}
    \label{fig:occl_vot2016}
\end{figure}

\renewcommand{\arraystretch}{0.9}
\begin{table*}[]
\begin{center}
\resizebox{\textwidth}{!}{
\begin{tabular}{clc|cccc|cccc}
\hline
  &                                       & \multicolumn{1}{c|}{} & \multicolumn{4}{c|}{\textbf{VOT 2014}}                                         & \multicolumn{4}{c}{\textbf{Imagenet Video}}                                       \\ \hline
  & Network Structure and Training Method & \textbf{Speed (FPS)} & \textbf{Accuracy} & \textbf{\# Drops} & \textbf{Robustness} & \textbf{Average} & \textbf{Accuracy} & \textbf{\# Drops} & \textbf{Robustness} & \textbf{Average}   \\
A & Feed Forward Network (GOTURN)~\cite{held}        & 168.77   & 0.61              & 35                  & 0.90             & 0.756            & 0.55               & 471                 & 0.95                & 0.750   \\
B & A + Imagenet Video Training           & 168.77              & 0.55              & 41                & 0.89                & 0.718            & 0.56              & 367               & 0.96                & 0.760               \\
C & One Layer LSTM                        & \textbf{213.27}     & 0.48              & 67                & 0.82                & 0.651            & 0.49              & 738               & 0.92                & 0.706               \\
D & C + Self-training                     & \textbf{213.27}     & 0.57              & 43                & 0.88                & 0.726            & 0.6               & 450               & 0.95                & 0.776               \\
E & D + Simulated Data                    & \textbf{213.27}     & 0.6               & 38                & 0.89                & 0.747            & 0.65              & 359               & 0.96                & 0.806               \\
F & E + Skip Layers                       & 160.72              & 0.62              & 29                & \textbf{0.92}       & 0.769            & 0.69              & 320               & \textbf{0.97}       & 0.828               \\
G & Full Model (F with two LSTM layers)   & 149.63              & 0.66              & 29                & \textbf{0.92}       & 0.789            & 0.68              & 257               & 0\textbf{.97}       & 0.826               \\
H & Full Model No ALOV                    & 149.63               & 0.6               & 28                & \textbf{0.92}       & 0.761            & \textbf{0.71}     & \textbf{233}      & \textbf{0.97}       & \textbf{0.842}      \\
I & Full Model No Imagenet Video          & 149.63              & 0.58              & 61                & 0.82                & 0.700            & 0.52              & 1096              & 0.88              & 0.700      \\
J & Full Model No LSTM Reset              & 149.63              & 0.54              & 47                & 0.87                & 0.705            & 0.61              & 539               & 0.94              & 0.775      \\
K & Full Model with GoogleNet~\cite{googlenet} conv layers & 77.29 & \textbf{0.68}     & \textbf{27}   & \textbf{0.92}       & \textbf{0.802}   & 0.69     & 274               & \textbf{0.97}       & 0.830                \\ \hline
\end{tabular}
}
\end{center}
\caption{Ablation Study. \textnormal{Average represents the arithmetic mean of accuracy and robustness, providing a single score to each method. Results on VOT 2014~\cite{vot2014} differ slightly from the VOT test suite, as they consider bounding boxes with a rotation angle, and we take the outermost points on these boxes as the ground truth labels.}}\vspace*{-8ex}
\label{table:ablation}
\end{table*}

\subsection{Robustness to Occlusion}
    We present two additional experiments showing that Re$^3$ performs comparatively well during occlusions. LSTMs can implicitly learn to handle occlusions because the structure of an LSTM can ignore information via the input and forget gates. This contrasts many other methods which assume all observations are useful, and may update their internal representation to include the occluder's appearance. In these experiments, we compare both the quality of track during occlusions as well as the difference in performance between overall scores and scores during occluded frames.

    First, we examine our performance on the VOT 2016 test set~\cite{vot2016}. Figure \ref{fig:occl_vot2016} shows the expected overlap measure of the same trackers from Figure \ref{fig:speed_plot} Right. Expected overlap represents the trackers' accuracy and robustness as a single number by performing many trials on subsets of the original data (more details available in~\cite{vot2015}). Each tracker has two points on the graph: the first for overall expected overlap, and the second for expected overlap during occlusion. Re$^3$ performs nearly as well as the top performers from VOT 2016~\cite{vot2016} however at a much higher frame rate, and outperforms many on occluded frames. Re$^3$'s performance degrades slightly during occlusions, but many of the other trackers drop in accuracy by more than 25\%. sKCF~\cite{skcf} also barely changes, and BDF~\cite{bdfAndMatFlow} actually improves during occlusions, however we outperform both of these methods on all frames and on occluded frames specifically.

    We also evaluate how Re$^3$'s performance degrades during occlusions across various IOU thresholds on OTB~\cite{otb}. Similar to the previous experiment, Figure \ref{fig:ope_tb50} compares the performance of trackers during all frames, and only during occluded frames. Again, we suffer a smaller loss in accuracy of 7.6 in relative percentage compared to other top methods (MUSTer~\cite{muster} 10.2\%, MEEM~\cite{meem} 8.3\%, STRUCK~\cite{hare2016struck} 16.1\%). The performance on both datasets under occlusion illustrate that our LSTM-based method offers significant robustness to occlusion - one of the most difficult challenges in object tracking.

\subsection{Ablation Study}
    Table \ref{table:ablation} examines how various changes to the network affect the speed and performance of Re$^3$ on the VOT 2014 and Imagenet Video test sets~\cite{vot2014, imagenetVideo}. The difference between model A and C is that model A has three fully-connected layers with 4096 outputs each, whereas C has one fully-connected layer with 2048 outputs, and one LSTM layer with 1024 outputs. Despite the small change, simply adding an LSTM to an existing tracker without any modification in the training procedure hinders performance. Self-training, learning to correct previous mistakes and prevent drift (model D), is clearly necessary when training a recurrent tracker. Other modifications tend to add slight improvements in both accuracy and robustness. At the expense of speed, we can attain even better results. Model K uses the GoogleNet~\cite{googlenet} architecture to embed the images, but is twice as slow. Model H, which was trained only on Imagenet Video~\cite{imagenetVideo} and simulated data, shows that by training on a fixed set of classes, performance improves on those classes but drops significantly on new objects (VOT 2014~\cite{vot2014}). Model I illustrates the need for a large training dataset, which seems especially important in terms of robustness. Model J shows the importance of resetting the LSTM state, as without the reset the network is much more affected by both parameter and model drift.

\subsection{Qualitative Results}
\begin{figure*}[!h]
    \centering
    \vspace{1ex}
    \resizebox{\textwidth}{!}{
    \includegraphics{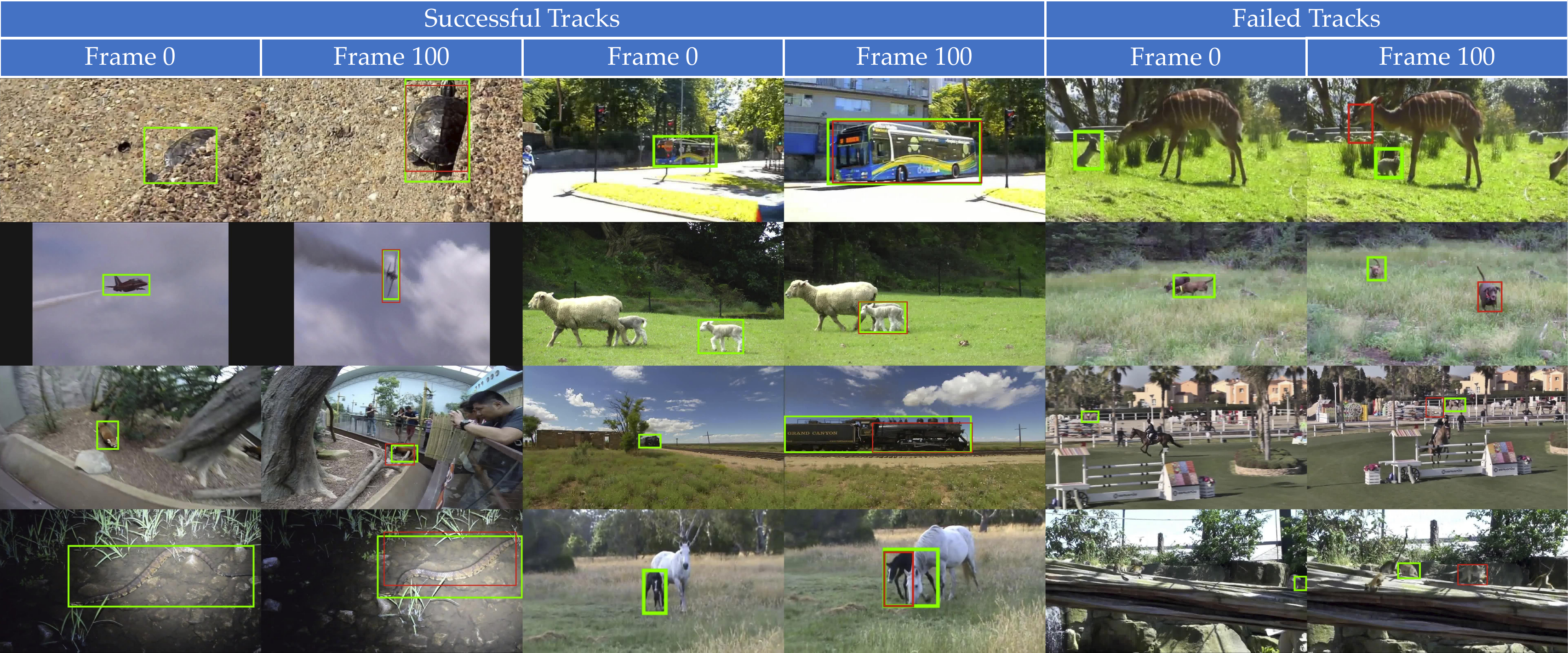}
    }\vspace*{-2ex}
    \caption{Qualitative Results: Examples of our tracker on the Imagenet Video~\cite{imagenetVideo} dataset. We show the initial frame as well as the 100th frame. Green represents the ground truth box and red represents Re$^3$'s output. We include challenging examples of scale, appearance, and aspect ratio change as well as occlusion. On the right, we show failures due to latching onto a foreground object, confusion with a similar object, latching onto an occluder, and small or ambiguous initial bounding box.}\vspace*{-2ex}
    \label{fig:qualitative}\vspace*{-2ex}
\end{figure*}

    We test our network on a variety of important and challenging never-before-seen domains in order to gauge Re$^3$'s usefulness to robotic applications. With a single initialization frame, our network performs remarkably well on challenging tasks such as shaky cellphone and car dashcam footage of a moving target, drone footage of multiple people simultaneously, and surveillance video of objects as small as $15 \times 20$ pixels. These results are shown in our supplemental video which can be found at \url{https://youtu.be/RByCiOLlxug}. We also include excerpts of our performance on challenging frames from Imagenet Video~\cite{imagenetVideo} in Figure \ref{fig:qualitative}.

\section{Conclusion}
    In this paper, we presented the first algorithm that uses a recurrent neural network to track generic objects in a variety of natural scenes and situations. Recurrent models offer a new, compelling method of tracking due to their ability to learn from many examples offline and  to quickly update online when tracking a specific object. Because they are end-to-end-trainable, recurrent networks can directly learn robustness to complex visual phenomena such as occlusion and appearance change. Our method demonstrates increased accuracy, robustness, and speed over comparable trackers, especially during occlusions. We showed how to efficiently and effectively train a recurrent network to learn from labeled videos and synthetic data. Ultimately we have shown that recurrent neural networks have great potential in the fast generic object tracking domain, and can be beneficial in robotics applications that need real-time performance on a limited computational budget.

\medskip

\bibliographystyle{ieee}
\bibliography{citations.bib}

\begin{thebibliography}{10}\itemsep=-1pt

\bibitem{tensorflow2015-whitepaper}
M.~Abadi et~al.
\newblock {TensorFlow}: Large-scale machine learning on heterogeneous systems,
  2015.
\newblock Software available from tensorflow.org.

\bibitem{arras2007using}
K.~O. Arras, O.~M. Mozos, and W.~Burgard.
\newblock Using boosted features for the detection of people in 2d range data.
\newblock In {\em IEEE International Conference on Robotics and Automation},
  pages 3402--3407. IEEE, 2007.

\bibitem{bengio2015scheduled}
S.~Bengio, O.~Vinyals, N.~Jaitly, and N.~Shazeer.
\newblock Scheduled sampling for sequence prediction with recurrent neural
  networks.
\newblock In {\em Advances in Neural Information Processing Systems}, pages
  1171--1179, 2015.

\bibitem{bertinetto2016fully}
L.~Bertinetto, J.~Valmadre, J.~F. Henriques, A.~Vedaldi, and P.~H. Torr.
\newblock Fully-convolutional siamese networks for object tracking.
\newblock In {\em European Conference on Computer Vision}, pages 850--865.
  Springer, 2016.

\bibitem{dsst}
M.~Danelljan, G.~H{\"a}ger, F.~Khan, and M.~Felsberg.
\newblock Accurate scale estimation for robust visual tracking.
\newblock In {\em British Machine Vision Conference}. BMVA Press, 2014.

\bibitem{danelljan2017discriminative}
M.~Danelljan, G.~H{\"a}ger, F.~S. Khan, and M.~Felsberg.
\newblock Discriminative scale space tracking.
\newblock {\em IEEE transactions on pattern analysis and machine intelligence},
  39(8):1561--1575, 2017.

\bibitem{danelljan2015convolutional}
M.~Danelljan, G.~Hager, F.~Shahbaz~Khan, and M.~Felsberg.
\newblock Convolutional features for correlation filter based visual tracking.
\newblock In {\em Proceedings of the IEEE International Conference on Computer
  Vision Workshops}, pages 58--66, 2015.

\bibitem{ccot}
M.~Danelljan, A.~Robinson, F.~S. Khan, and M.~Felsberg.
\newblock Beyond correlation filters: Learning continuous convolution operators
  for visual tracking.
\newblock In {\em European Conference on Computer Vision}, pages 472--488.
  Springer, 2016.

\bibitem{kahou2017ratm}
S.~Ebrahimi~Kahou, V.~Michalski, R.~Memisevic, C.~Pal, and P.~Vincent.
\newblock Ratm: Recurrent attentive tracking model.
\newblock In {\em Proceedings of the IEEE Conference on Computer Vision and
  Pattern Recognition Workshops}, pages 10--19, 2017.

\bibitem{edoardo2014matrioska}
M.~Edoardo~Maresca and A.~Petrosino.
\newblock The matrioska tracking algorithm on ltdt2014 dataset.
\newblock In {\em Proceedings of the IEEE Conference on Computer Vision and
  Pattern Recognition Workshops}, pages 706--711, 2014.

\bibitem{fragkiadaki2015recurrent}
K.~Fragkiadaki, S.~Levine, P.~Felsen, and J.~Malik.
\newblock Recurrent network models for human dynamics.
\newblock In {\em Proceedings of the IEEE International Conference on Computer
  Vision}, pages 4346--4354, 2015.

\bibitem{gan2015first}
Q.~Gan, Q.~Guo, Z.~Zhang, and K.~Cho.
\newblock First step toward model-free, anonymous object tracking with
  recurrent neural networks.
\newblock {\em arXiv preprint arXiv:1511.06425}, 2015.

\bibitem{kitti}
A.~Geiger, P.~Lenz, and R.~Urtasun.
\newblock Are we ready for autonomous driving? the kitti vision benchmark
  suite.
\newblock In {\em Proceedings of the IEEE Conference on Computer Vision and
  Pattern Recognition}, 2012.

\bibitem{dsstCode}
{Georg Nebehay}.
\newblock Dsst: Discriminative scale space tracker.
\newblock \url{https://github.com/gnebehay/DSST}.

\bibitem{greff2016lstm}
K.~Greff, R.~K. Srivastava, J.~Koutn{\'\i}k, B.~R. Steunebrink, and
  J.~Schmidhuber.
\newblock Lstm: A search space odyssey.
\newblock {\em IEEE transactions on neural networks and learning systems},
  2016.

\bibitem{hare2016struck}
S.~Hare, S.~Golodetz, A.~Saffari, V.~Vineet, M.-M. Cheng, S.~L. Hicks, and
  P.~H. Torr.
\newblock Struck: Structured output tracking with kernels.
\newblock {\em IEEE transactions on pattern analysis and machine intelligence},
  38(10):2096--2109, 2016.

\bibitem{prelu}
K.~He, X.~Zhang, S.~Ren, and J.~Sun.
\newblock Delving deep into rectifiers: Surpassing human-level performance on
  imagenet classification.
\newblock In {\em Proceedings of the IEEE International Conference on Computer
  Vision}, pages 1026--1034, 2015.

\bibitem{held}
D.~Held, S.~Thrun, and S.~Savarese.
\newblock Learning to track at 100 fps with deep regression networks.
\newblock In {\em European Conference on Computer Vision}, pages 749--765.
  Springer, 2016.

\bibitem{kcf}
J.~F. Henriques, R.~Caseiro, P.~Martins, and J.~Batista.
\newblock High-speed tracking with kernelized correlation filters.
\newblock {\em IEEE Transactions on Pattern Analysis and Machine Intelligence},
  37(3):583--596, 2015.

\bibitem{muster}
Z.~Hong, Z.~Chen, C.~Wang, X.~Mei, D.~Prokhorov, and D.~Tao.
\newblock Multi-store tracker (muster): A cognitive psychology inspired
  approach to object tracking.
\newblock In {\em Proceedings of the IEEE Conference on Computer Vision and
  Pattern Recognition}, pages 749--758, 2015.

\bibitem{kang2017t}
K.~Kang et~al.
\newblock T-cnn: Tubelets with convolutional neural networks for object
  detection from videos.
\newblock {\em IEEE Transactions on Circuits and Systems for Video Technology},
  2017.

\bibitem{kingma2014adam}
D.~Kingma and J.~Ba.
\newblock Adam: A method for stochastic optimization.
\newblock {\em arXiv preprint arXiv:1412.6980}, 2014.

\bibitem{vot2014}
M.~Kristan et~al.
\newblock The visual object tracking vot2014 challenge results.
\newblock 2014.

\bibitem{vot2015}
M.~Kristan et~al.
\newblock The visual object tracking vot2015 challenge results.
\newblock In {\em Proceedings of the IEEE International Conference on Computer
  Vision Workshops}, pages 1--23, 2015.

\bibitem{vot2016}
M.~Kristan et~al.
\newblock The visual object tracking vot2016 challenge results.
\newblock In {\em European Conference on Computer Vision Workshops}, 2016.

\bibitem{LanRSS17}
Z.~Lan, M.~Shridhar, D.~Hsu, and S.~Zhao.
\newblock Xpose: Reinventing user interaction with flying cameras.
\newblock In {\em Robotics: Science and Systems}, 2017.

\bibitem{tsne}
L.~v.~d. Maaten and G.~Hinton.
\newblock Visualizing data using t-sne.
\newblock {\em Journal of Machine Learning Research}, 9(Nov):2579--2605, 2008.

\bibitem{bdfAndMatFlow}
M.~E. Maresca and A.~Petrosino.
\newblock Clustering local motion estimates for robust and efficient object
  tracking.
\newblock In {\em European Conference on Computer Vision}, pages 244--253.
  Springer, 2014.

\bibitem{nam2016mdnet}
H.~Nam and B.~Han.
\newblock Learning multi-domain convolutional neural networks for visual
  tracking.
\newblock In {\em Proceedings of the IEEE Conference on Computer Vision and
  Pattern Recognition}, pages 4293--4302, 2016.

\bibitem{ondruska2016deep}
P.~Ondruska and I.~Posner.
\newblock Deep tracking: Seeing beyond seeing using recurrent neural networks.
\newblock In {\em Thirtieth AAAI Conference on Artificial Intelligence}, 2016.

\bibitem{premebida2007lidar}
C.~Premebida, G.~Monteiro, U.~Nunes, and P.~Peixoto.
\newblock A lidar and vision-based approach for pedestrian and vehicle
  detection and tracking.
\newblock In {\em Intelligent Transportation Systems Conference, 2007. ITSC
  2007. IEEE}, pages 1044--1049. IEEE, 2007.

\bibitem{richter2016playing}
S.~R. Richter, V.~Vineet, S.~Roth, and V.~Koltun.
\newblock Playing for data: Ground truth from computer games.
\newblock In {\em European Conference on Computer Vision}, pages 102--118.
  Springer, 2016.

\bibitem{imagenetVideo}
O.~Russakovsky et~al.
\newblock Imagenet large scale visual recognition challenge.
\newblock {\em International Journal of Computer Vision}, 115(3):211--252,
  2015.

\bibitem{dart}
T.~Schmidt, R.~A. Newcombe, and D.~Fox.
\newblock Dart: Dense articulated real-time tracking.
\newblock In {\em Robotics: Science and Systems}, 2014.

\bibitem{alov}
A.~W. Smeulders, D.~M. Chu, R.~Cucchiara, S.~Calderara, A.~Dehghan, and
  M.~Shah.
\newblock Visual tracking: An experimental survey.
\newblock {\em IEEE Transactions on Pattern Analysis and Machine Intelligence},
  36(7):1442--1468, 2014.

\bibitem{skcf}
A.~Solis~Montero, J.~Lang, and R.~Laganiere.
\newblock Scalable kernel correlation filter with sparse feature integration.
\newblock In {\em Proceedings of the IEEE International Conference on Computer
  Vision Workshops}, pages 24--31, 2015.

\bibitem{googlenet}
C.~Szegedy et~al.
\newblock Going deeper with convolutions.
\newblock In {\em Proceedings of the IEEE Conference on Computer Vision and
  Pattern Recognition}, pages 1--9, 2015.

\bibitem{vinyals2015show}
O.~Vinyals, A.~Toshev, S.~Bengio, and D.~Erhan.
\newblock Show and tell: A neural image caption generator.
\newblock In {\em Proceedings of the IEEE Conference on Computer Vision and
  Pattern Recognition}, pages 3156--3164, 2015.

\bibitem{asmsCode}
T.~Vojir.
\newblock Robust scale-adaptive mean-shift for tracking.
\newblock \url{https://github.com/vojirt/asms}.

\bibitem{kcfCode}
T.~Vojir.
\newblock Tracking with kernelized correlation filters.
\newblock \url{https://github.com/vojirt/kcf}.

\bibitem{asms}
T.~Vojir, J.~Noskova, and J.~Matas.
\newblock Robust scale-adaptive mean-shift for tracking.
\newblock In {\em Scandinavian Conference on Image Analysis}, pages 652--663.
  Springer, 2013.

\bibitem{otb}
Y.~Wu, J.~Lim, and M.-H. Yang.
\newblock Online object tracking: A benchmark.
\newblock In {\em Proceedings of the IEEE conference on Computer Vision and
  Pattern Recognition}, pages 2411--2418, 2013.

\bibitem{zeiler2014visualizing}
M.~D. Zeiler and R.~Fergus.
\newblock Visualizing and understanding convolutional networks.
\newblock In {\em European conference on computer vision}, pages 818--833.
  Springer, 2014.

\bibitem{meem}
J.~Zhang, S.~Ma, and S.~Sclaroff.
\newblock Meem: robust tracking via multiple experts using entropy
  minimization.
\newblock In {\em European Conference on Computer Vision}, pages 188--203.
  Springer, 2014.

\end{thebibliography}
\nocite{*}

\end{document}